\documentclass{article}
\usepackage{graphics} 
\usepackage{amsmath} 
\usepackage{amssymb}  
\usepackage{comment}
\usepackage{url}
\usepackage{graphicx}
\usepackage{subcaption}
\usepackage{wrapfig}
\usepackage{xcolor}
\usepackage{makecell}

\usepackage[ruled,vlined]{algorithm2e}
\definecolor{blue}{RGB}{0, 0, 255}
\usepackage[preprint]{corl_2020}
\usepackage[colorinlistoftodos]{todonotes}

\usepackage{xcolor}
\definecolor{darkgreen}{rgb}{0.0, 0.5, 0.0}
\definecolor{orange0}{rgb}{0.9, 0.45, 0.2}

\newcommand{\mycomment}[1] {}

\title{ACNMP: Skill Transfer and Task Extrapolation through Learning from Demonstration and Reinforcement Learning via Representation Sharing}

\author{
  M. Tuluhan Akbulut\thanks{Corresponding author: \texttt{tuluhan.akbulut@boun.edu.tr}}\\
  Bogazici University,
  Turkey
  \And
  Erhan Oztop\\
  Ozyegin University, Turkey  
  \And
  M. Yunus Seker\\
  Bogazici University, 
  Turkey
  \And
  Honghu Xue\\
  University of Luebeck,
  Germany
  \And
  Ahmet E. Tekden\\
  Bogazici University, 
  Turkey
  \And
  Emre Ugur\\
  Bogazici University, 
  Turkey
  
}

\begin{document}
\maketitle

\vspace{-0.6cm}
\begin{abstract}
To equip  robots with dexterous skills, an effective approach is to first transfer the desired skill via Learning from Demonstration (LfD), then let the robot improve it by self-exploration via Reinforcement Learning (RL). In this paper, we propose a novel LfD+RL framework, namely Adaptive Conditional Neural Movement Primitives (ACNMP), that allows efficient policy improvement in novel environments and effective skill transfer between different agents. This is achieved through exploiting the latent representation learned by the underlying Conditional Neural Process (CNP) model, and simultaneous training of the model with supervised learning (SL) for acquiring the demonstrated trajectories and via RL for new trajectory discovery. Through simulation experiments, we show that (i) ACNMP enables the system to extrapolate to situations where pure LfD fails; (ii) Simultaneous training of the system through SL and RL  preserves the shape of demonstrations while adapting to novel situations due to the shared representations used by both learners; (iii) ACNMP enables order-of-magnitude sample-efficient RL in extrapolation of reaching tasks compared to the existing approaches; (iv) ACNMPs can be used to implement skill transfer between robots having different morphology, with competitive learning speeds and importantly with less number of assumptions compared to the state-of-the-art approaches. Finally, we show the real-world suitability of ACNMPs through real robot experiments that involve obstacle avoidance, pick and place and pouring actions.
\end{abstract}



\vspace{-.2cm}
\section{Introduction}
  \addtocounter{footnote}{-1}%
\footnotetext{Code is available at:\url{https://mtuluhanakbulut.github.io/ACNMP}}
\label{intro}
\vspace{-.2cm}To equip  robots with dexterous skills, an effective approach is to first transfer an approximate version of the desired skill, then let the robot improve it by self-exploration. For the initial transfer of the skill, learning from demonstration (LfD) is a natural choice, where the robot is provided with a set of expert movement demonstrations, by which it learns and reproduces the  demonstrations \cite{Argall2009}. If the demonstrations are obtained in the intrinsic coordinates of the robot, a simple playback controller, or a neural network capturing the state to action mapping can synthesize the initial skill (e.g. \citep{Peternel2014, babic_hale_oztop2011}).
To improve the quality of the initial skill obtained, and/or to adapt it to novel environments, the movement policy of the robot is typically modified via  Reinforcement Learning (RL)  \cite{sutton2018reinforcement},  where the reward function defined by the designer determines how the initial skill is to be modified (e.g. \cite{Kober2013}). 

Instead of targeting a single task, recent efforts in LfD+RL approaches focus on developing methods to allow the representation of multiple policies that can be flexibly adapted according to the goals and contexts, while retaining original skills if needed \citep{osa2018}. We believe that this is one of the keys for paving the road for life-long learning. We consider the following three challenges in this context:

\textit{Sample-efficiency.} The relations between task parameters and the required policy are often highly non-linear and complex in real-world problems. Consequently, excessive policy updates may be required to reach a policy to make the robot perform in new contexts. Therefore, sample-efficient robust machine learning methods are needed to adapt the initial policies to the novel ones \cite{dulac2019challenges}. 

\textit{Avoiding interference of novel goals with the initial demonstrations}. The skill demonstrations provided by humans include intrinsic features that are difficult to encode in a reward function.
These characteristics might be related to safety, robot and environment constraints, or other user-preferences. Thus, the second challenge is to update the policy towards the fulfillment of novel contexts/goals while maintaining the characteristics of the initial demonstration. 

\textit{Skill transfer between agents with different embodiments}. When the skill to be transferred is not represented in the robot intrinsic coordinates then the skill transfer requires the additional step of converting the demonstration to the robot intrinsic coordinates. This is problematic if the mapping is not well defined or multiple solutions exist, in which case, it is desirable to  constraint the mapping by the space of demonstrations. Therefore, an important challenge in transferring skills between agents is to develop mechanisms that enable such transfer automatically without explicitly designing the coordinate between agents.

In this study, we address the aforementioned challenges by: {\bf (i)} Developing an LfD+RL framework that allows the robot to learn complex relations between the task parameters and generated motions in a sample-efficient way. {\bf (ii)} Ensuring the LfD+RL framework to adapt the policy to new environments while reflecting and maintaining skill knowledge captured from the initial demonstrations. {\bf (iii)} Creating a common representation space between robots with different embodiments, which allows automatic skill transfer between them.

To address these challenges, we leverage the recent advances in deep learning, particularly in Conditional Neural Processes\cite{CNP}, and extend a previously formulated LfD framework\cite{sekerconditional} with a novel RL component. Our proposed LfD+RL framework, 
namely Adaptive Conditional Neural Movement Primitives (ACNMP)  learns the distribution of  the input movement trajectories conditioned on the user set task  parameters by representing the  relations  between  the task parameters and  movement  trajectories  from a few  demonstrations. When the system is queried outside of the range it was trained for, the generated trajectories may fail to achieve the desired goals.  This may also happen when the system is placed in a new environment.  In these cases, the system starts to adapt its skill representation through simultaneous RL and LfD learning. To be concrete, the system updates its internal parameters via RL guided actions, which is alternated with  error-based supervised learning (SL) based on the demonstration set. 
An ACNMP system employs an encoder-decoder network with a powerful latent space representation (i.e. a CNMP system \cite{sekerconditional})  for implementing the LfD part. In ACNMP, this network, in a novel way, is also given to  implement the policy network for the RL part. The ACNMP architecture not only allows acquiring an ever growing set of movement trajectories but also facilitates skill transfer between two morphologically different robots. To do this, two ACNMP models are trained to develop a common representation in their latent layers by using a proxy skill demonstration that is available to both. Then one robot can automatically learn a novel, possibly more complex, task by observing its execution by its peer. 

We conducted simulation experiments to justify the aforementioned features of ACNMP and show its superiority over the existing methods. Moreover, through real robot experiments, we showed the suitability of ACNMP for real-world deployment.
\vspace{-.25cm}
\section{Related Work}
\label{sec:relatedworks}
\vspace{-.25cm}
{\bf Learning Movement Primitives from Demonstrations}
LfD \citep{Argall2009} has been extensively used in robotic problems including object grasping and manipulation \citep{Calinon2009,Asfour2008,Ben2012,Pastor2011,Muhlig2012}. Among others, learning methods that are based on dynamic systems \citep{Schaal2006}, statistical modeling \citep{Calinon2016} and their combination \citep{Girgin2018,Ugur2020} have been popular in recent years. Dynamic Movement Primitives (DMPs) encode the demonstrated trajectory as a set of differential equations, implementing a spring-mass-damper system extended with a non-linear function.  Encountered with novel situations, the model parameters of the DMP can be adjusted using RL \citep{Theodorou2010,Colome2014}. 
While DMPs generate deterministic trajectories, Probabilistic Movement Primitives (ProMPs) \citep{Paraschos2013} can encode a distribution of trajectories and generate stochastic policies. CNMPs \citep{sekerconditional} can also encode trajectory distributions while learning non-linear relationships. We select CNMPs as our LfD model and we will compare the performance and generalization capabilities of our method (ACNMP) against adaptive ProMPs in this paper. 


{\bf Adaptation of Primitives}
Sample efficiency is critical for RL when applied in the real-world, henceforth, LfD is often combined with RL to facilitate sample-efficient learning~\citep{dulac2019challenges}. These approaches can be viewed as two complementary categories. The first one employ an RL engine at the core and use demonstrations to help RL by providing expert input \citep{hester2018deep, vecerik2019practical,rajeswaran2018learning,zhu2018reinforcement}. 
For example, in \citep{hester2018deep}, consistency between the demonstrated and self-acquired trajectories is maintained by defining a trade-off cost function that takes into account expert knowledge in integrating the reward obtained from the environment. However, unless the trade-off weighting hyper-parameters are carefully tuned, the agent may get stuck with the performance of the demonstrator or fail to benefit from the demonstration \citep{li2018learning}. 
The second category, which also includes our work, takes an LfD engine as the core learner and improves its performance by extending it with RL. Here, our approach is closely related to \citep{Stark2019,Ewerton2019}, where ProMPs are used to encode demonstrations and they adapt ProMPs to new task constraints via RL. In \citep{Stark2019}, a planar robot arm learns separate ProMPs for pushing objects to different goal points through demonstrations, and extends this skill to novel goal points by exploiting the previously learned ProMP models. The KL divergence is used to stay close to previous parameters of the model, avoiding distorting the shape of the movement \citep{peters2010}. In our study, the same skill with different task parameters (such as goal points) is encoded in a single ACNMP model. Similar to ours, \citep{Ewerton2019} also uses a single model which combines ProMP and Gaussian Processes to encode a parametric skill and condition it with the corresponding task parameters. For adaptation, RL is used to find the relations between the task parameters and the ProMP model parameters by using a trajectory relevance metric. Compared to the above studies, our model does not require explicit optimization using metrics such as additional/weighted loss, relevance or KL-divergence. Instead, ACNMP is trained together with the demonstrated trajectories and the newly explored ones, automatically preserving the old skills while extending the model to the new task parameters thanks to the robust and flexible generated representations. The idea of augmenting the demonstrations in online learning settings is also studied in \citep{ross2011reduction}, where only supervised learning is employed.

Transfer learning has also been an important challenge in robotics \citep{Taylor2009}. The correspondences of states between different agents are established by considering all state pairs \citep{Taylor2008}, by manually designing a common feature space \citep{Ammar2012}, or by aligning states via unsupervised manifold alignment \citep{Ammar2015}. \citet{gupta2017learning} aligns the states using expectation-maximization based dynamic time warping, and then found a common feature space using non-linear embedding functions. Different from the previous studies, our method does not require an  explicit state matching step. It forms a common feature space that encodes the correspondence of the trajectories rather than single states.
\vspace{-.25cm}\section{Method}
\label{Method}
\begin{figure*}[t]
    \centering
    \includegraphics[width=1.0\linewidth]{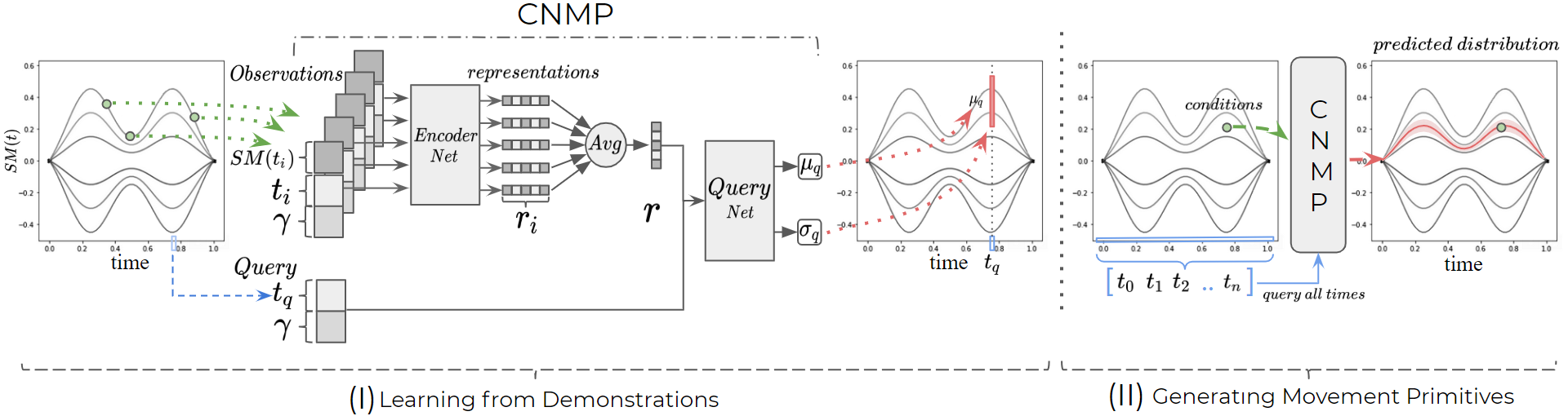}
    \caption{Policy network training and trajectory generation during LfD. See Section \ref{Method}.A for details.}\vspace{-.6cm}
    \label{fig:cnmp}
\end{figure*}
\vspace{-.25cm}As mentioned in the introduction, our system is composed of LfD and RL components. The {\bf LfD component}  of our system is composed of an encoder-decoder network, namely CNMP \cite{sekerconditional}. The encoder layer of the CNMP learns a representation of trajectory points and other variables conditioned on the time. The decoder layer takes the learned representations and outputs the mean and  variance of a Gaussian distribution as a function of time. 
Figure~\ref{fig:cnmp}.I shows the training procedure of a CNMP using a hypothetical 1D scenario. In each training iteration, a changing number of random sensorimotor time and value pairs, named observation points, are sampled from a random demonstration trajectory. The observation points are passed through the parameter sharing encoder network in order to be transformed into their corresponding latent space representations, which are then merged into one single general representation. The produced general representation is concatenated with the target query time value and passed through the decoder network that produces a mean and a variance, which corresponds to the sensorimotor value distribution of the query time based on the sampled observations. The network is trained end-to-end with stochastic gradient descent algorithm via the loss function: $\mathcal{L}(\theta,\phi) = -\log P(SM(t_q) \mid \mu_q,\mathop{\textrm{softplus}}(\sigma_q))$ 
where $\mu_q$ and $\sigma_q$ are the predicted distribution parameters over the queried time-step $t_q$, and $SM(t_q)$ is the recorded sensorimotor value at the queried time.  Conditioning on task parameters ($\gamma$) is achieved by concatenating the parameters with the time points. After training, the CNMP can be queried to produce a trajectory that matches a desired set of outputs at desired time-points (Fig.~\ref{fig:cnmp}.II).

The {\bf RL component} of our system follows a policy gradient RL approach, where the behavior of an RL agent with stochastic policy is governed by the policy $\pi_\theta(a,s)=P(a_t \in A| s_t \in S , \theta)$, where $S$ and $A$ denote state and action spaces, and $\theta$ indicates the parameters of a function approximator. We define the policy to depend on context $c$, which is composed of time and possibly task parameters. 
Thus we have $\pi_\theta(a,s,c)$ as our policy.
As our agent needs to generate continuous actions,  we adopt a policy-gradient-based algorithm for learning.  Policy gradient algorithms aim to maximize the expected total reward {\small $R(\tau)= \sum\limits_{t=1,..T} R(s_t,a_t,c_t)$} over a state-action trajectory $\tau = \left \{s_0,a_0,...s_T,a_T\right \}$ with respect to the policy parameters $\theta$ \citep{Kober2014IT}. In our setting, the update on policy parameters $\theta$  becomes: 
\begin{equation} \label{eq:objective_simplified}
\nabla_{\theta} J(\theta)=\mathbb{E}_{\pi_\theta}[\sum\limits_{t=1}^T \nabla_{\theta} \log{\pi_\theta(s_t, a_t, c_t)}(\sum\limits_{t'=t}^T R(s_{t'} ,a_{t'}, c_{t'}))]
\end{equation}
\begin{figure*}[t]
    \centering
    \includegraphics[width=0.85
    \linewidth]{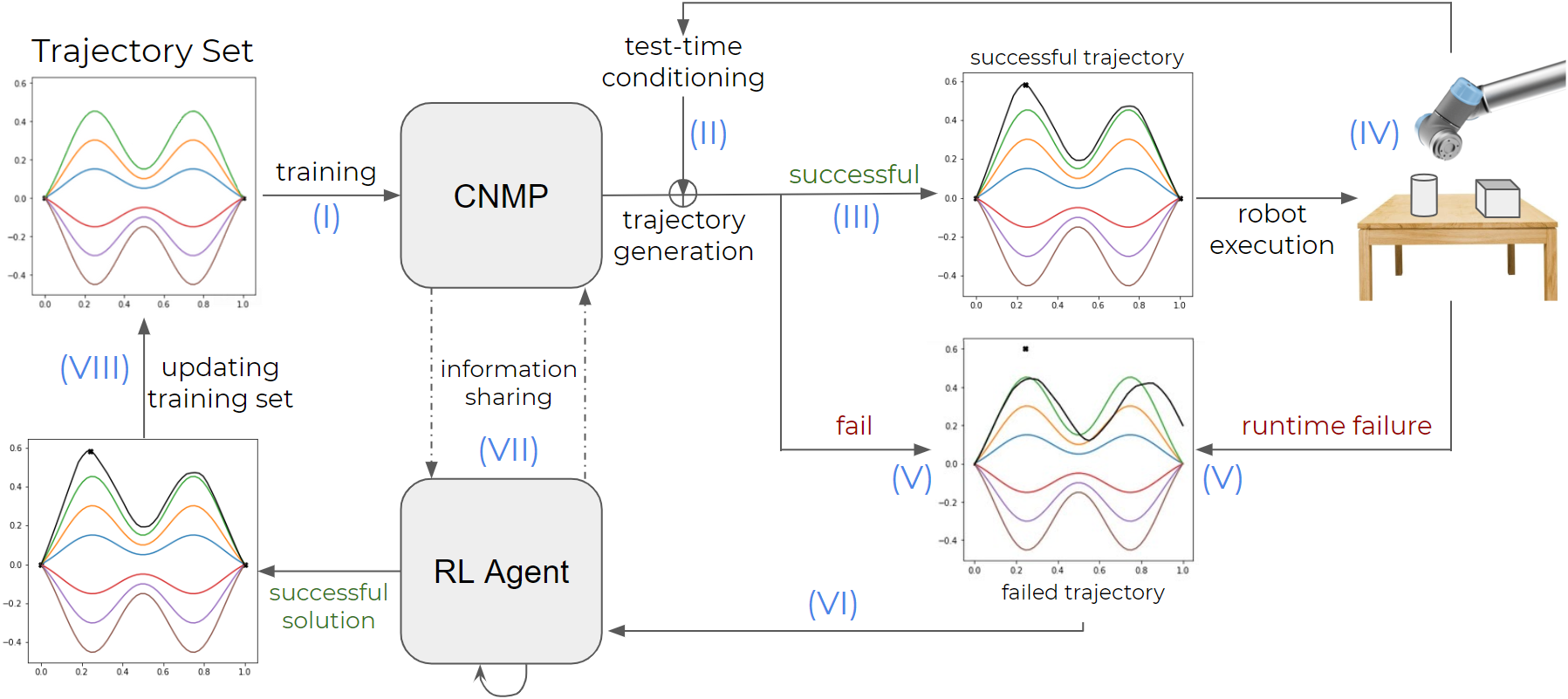}
    \caption{The proposed ACNMP architecture and the learning loop.}\vspace{-.5cm}
    \label{fig:acnmp}
\end{figure*}
Fig.~\ref{fig:acnmp} provides the general proposed LfD+RL framework. Given the demonstration trajectories, the base CNMP model is trained (I) following the steps outlined in the first paragraph. In order to achieve the goal with given task parameters, CNMP model is conditioned with the task parameters (II) and generates the required trajectory (III), which is executed by the robot (IV). In case a failure is observed in extrapolation cases (V), RL agent steps in (VI) and updates the CNMP model via policy gradient approach ; while the same model continues being trained using demonstration trajectories (VII). After the RL agent finds the optimal trajectory for the new task , this new trajectory is regarded as a new demonstration and fed back to the LfD system (VIII).

{\bf Simultaneous Training with SL and RL:} A crucial aspect of our framework is that during the application of RL updates to discover a suitable trajectory, learning based on expert trajectories is simultaneously carried out to preserve the learned knowledge. This procedure ensures the formation of a model that not only produces novel trajectories that satisfy the new constraints but also  reproduces previously learned trajectories. In practice, this allows the system to generate novel trajectories that share similar characteristics with the demonstrated ones. We train the neural network using two loss functions for corresponding objectives, \ref{eq:objective_simplified} and the maximum likely-hood loss function of CNMP to achieve the aforementioned effect. For the inputs (observation and task parameters) coming from demonstrations, supervised learning loss is used to update the network. When the inputs are novel (coming from extrapolation cases), the same network is updated by policy gradient loss. 
If we manipulate the CNMP loss in equation 1, it can be seen that it is similar to policy gradient loss: 
\begin{equation} \label{eq:objective2}
\nabla_{\theta} J(\theta)=\mathbb{E}_{\textrm{CNMP}_{\theta}}[(\sum\limits_{t=1}^T \nabla_{\theta} \log_{\textrm{CNMP}_{\theta}}(a_{t, \textrm{demo}}, c_t))]
\end{equation} 
where $R(\tau)= 1$ as the true action ($a_{t,\textrm{demo}}$)  is known. This manipulation can be thought as follows: policy gradient adjusts policy distribution to make the most rewarding action most likely whereas CNMP adjusts trajectory distribution to make the observed actions most likely. Note that the imbalance in the number of initial demonstrations and the new trajectories does not hurt the target distribution since CNP can represent multimodal distributions \cite{CNP}.

{\bf Assimilating New RL Solution into the CNMP:} In finding the new solution, RL agent only maximizes the reward while using only conditioning points as observations. However, CNMP is trained by sampling random observations from the trajectory. Therefore, the solution proposed by RL is added to the trajectory dataset and CNMP is trained further as described at the beginning of this section \ref{Method}. This assimilation step ensures the capability to condition CNMP from any point in the range of previous expert demonstrations and newly found RL solution.

{\bf Transfer Learning:}
To transfer knowledge between different agents, two agents initially learn skills with their own ACNMPs that are trained together for the proxy tasks to find common latent representations from demonstrations. Then, the target agent uses these representations with RL to learn the test task demonstrated by the source agent. Note that the problem of matching  actions executed by robots with a different number of joints and/or sampled with different frequencies is naturally handled by the proposed system as whole trajectories are mapped to the same length latent vectors. This mapping is achieved by requiring the latent spaces of two CNMPs to be as close as possible by including a latent space distance term in the loss function:
\begin{equation} \label{eq:losstransferFunction}
    \mathcal{L}(\theta_1,\phi_1,\theta_2,\phi_2) = \mathcal{L_{\textrm{CNMP1}}}(\theta_1,\phi_1) + \mathcal{L_{\textrm{CNMP2}}}(\theta_2,\phi_2) + \mathcal{L_{\textrm{MSE}}}(R_1, R_2) 
\end{equation} 
where $\mathcal{L_{\textrm{MSE}}}$ stands for mean squared error, $R_1$ denotes representation coming from encoder of first CNMP and $R_2$ denotes representation coming from encoder of the second CNMP.
After the latent space is formed, the knowledge transfer scenario folds out as follows. It is assumed that the first robot knows how to solve the test task; yet, the second robot has no knowledge of the task, and it would need many iterations to learn to solve the task. In this case, by looking at the solution trajectory of the first robot, the latent space representation can be found by the encoder of the first CNMP. Giving this representation to the decoder of the second CNMP results in the interpretation of the first robot's solution in terms of the second robot's joint angles. Afterwards, the RL agent can optimize the trajectory quickly to solve the task. 

\vspace{-.2cm}\section{Experimental Results}
\vspace{-.1cm}
\label{sec:result}
The performance of our system is evaluated in four simulations and two real robot experiments. For two experiments with which we compare our methods against the source code was not available, so the results pertaining those are obtained with our implementation based on the algorithms (including reward functions) given in \citet{Stark2019,gupta2017learning}. The implementation details can be found in the Appendix \ref{implementation_details}.


\vspace{-.2cm}
\subsection{Extrapolation in 2D Obstacle Avoidance}
\begin{figure}
\centering
    \includegraphics[width=0.9\linewidth]{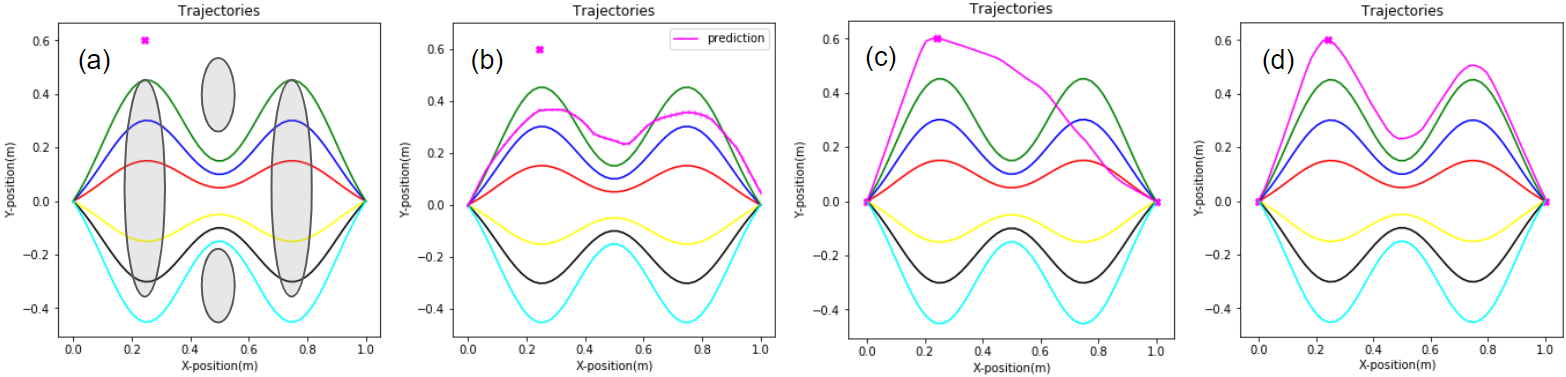}
    \caption{(a) The simulated 2d obstacle avoidance task. The 6 colored trajectories correspond to the demonstrated trajectories, the gray ellipses represent the underlying hypothetical obstacles and the pink red point shows the conditioned point. (b) LfD model fails to extrapolate. (c) Only RL-based LfD trajectory. (d) Simultaneous RL \& LfD-based ACNMP. }
    \label{fig:exp1}\vspace{-.3cm}
\end{figure}

\vspace{-.1cm}
This section aims to first verify that the original CNMP formulation fails to generalize when conditioned with points outside the learning range and then to evaluate the effectiveness of our proposed method. For this purpose, we simulated an obstacle avoidance task in a 2D world where different trajectories are given as demonstrations that avoid static obstacles. The six demonstrations that start from the same position and end at the same position are shown in Fig.~\ref{fig:exp1}(a) with different colors. Given a different via-point (shown with a dot), a trajectory whose shape is similar to the demonstrated ones is expected to be generated. However, conditioned with a dot outside of the training range, LfD model generates a disparate trajectory that does not pass through the conditioned point (Fig.~\ref{fig:exp1}(b)). ACNMP adapts network weights using previous demonstrations via supervised learning and new trajectories generated by RL setting the reward as the minus distance between the generated and the target points. After 35 roll-outs, ACNMP generates a trajectory passing through the given purple-dots while preserving the shape similar to initial demonstrations (Fig.~\ref{fig:exp1}(d)). Note that, when only RL is applied on the learned LfD model, the system can also generate a trajectory passing through the required point, but without preserving the shape  (Fig.~\ref{fig:exp1}(c)). 
We further analyzed our ACNMP system in extreme extrapolation cases and observed that while the constraints are satisfied, the shapes of the trajectories diverge from the demonstrated ones (see Appendix \ref{extreme_extrapolation} for details).

\begin{figure}
\centering
    \includegraphics[width=0.9\linewidth]{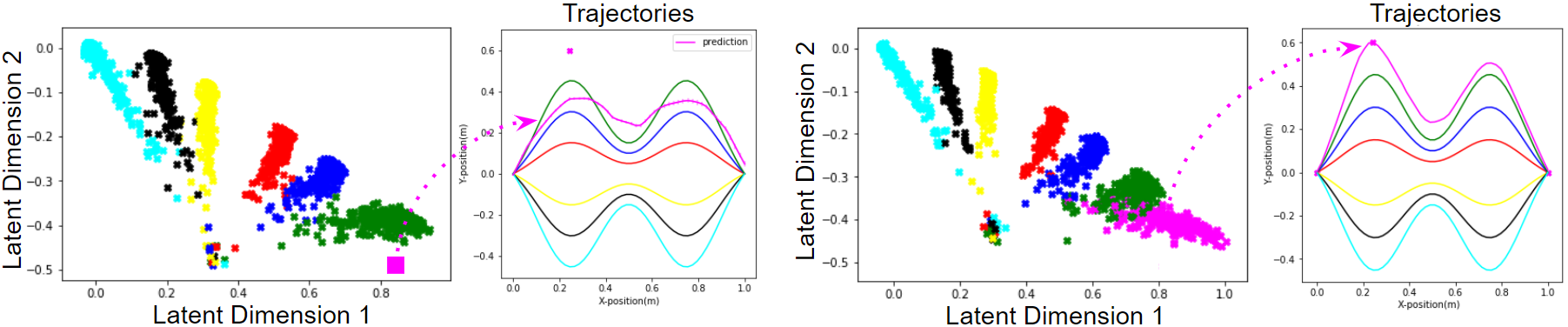}
    \caption{Illustration of the representations for each trajectory in the latent representation space. The left and right parts show latent representation before and after RL solution is assimilated into the model, respectively.}
    \label{fig:latent}\vspace{-.6cm}
\end{figure}

{\it Assimilating new constraints into CNMP representations:}
We investigate how different demonstrations are encoded in the  representation layer of the neural network and how this representation changes after ACNMP learning. For this purpose, points from the trajectories are sampled and the corresponding activations in the 2-neuron latent representation layer are visualized in Fig.~\ref{fig:latent}. As shown, each separate trajectory is represented in a different cluster in the latent space. After RL is applied and the newfound trajectory is integrated into the CNMP, it is represented in a separate cluster in the latent space. Note that, if the same activation were used in the latent space before the RL step, the reproduced trajectory could not satisfy the constraints (as shown in the left part of figure \ref{fig:latent}). This analysis shows that our method enables the robust integration of new constraints into the existing representations. 



%
\vspace{-0.25cm}
\subsection{Adaptation to Novel Environments in Simulation Experiments}
\begin{figure}[t]
\centering
\includegraphics[width=\linewidth]{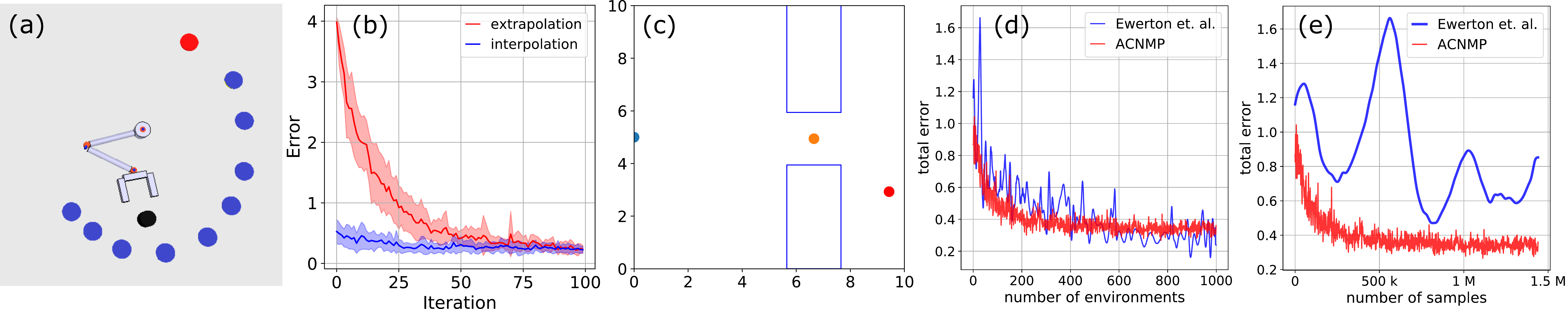}

    \caption{Tasks and performance comparisons for \citet{Stark2019} (a-b) and \citet{Ewerton2019} (c-e)}\vspace{-.5cm}
    \label{fig:comparisonpromp}
\end{figure}

\vspace{-0.15cm}
We evaluate the extrapolation capabilities of ACNMP in a simulated pushing and obstacle avoidance tasks, and compare the results with the state-of-the-art approaches \citep{Stark2019,Ewerton2019} in two experiments. In the first experiment, we replicated the experimental setup of \cite{Stark2019} in CoppeliaSim \citep{coppeliaSim}.
In this task, the task of a joint-controlled 3-dof planar robot arm is to push a cylindrical object from a fixed position to one of 10 different target positions. The robot is provided with demonstrations for 9 different target points and is expected to generate a trajectory and adapt if necessary to push the object to the remaining target point (Fig.~\ref{fig:comparisonpromp} (a)). Both \cite{Stark2019} and us set the reward to minus distance between the generated and target positions of the object. In \citep{Stark2019}, pushing the cylinder to each target was represented by a single ProMP model, whereas the joint trajectories were conditioned on target position in our model, therefore all skills were encoded into a single ACNMP. We analyzed two different situations, namely interpolation and extrapolation cases, separately and provided the performance of ACNMP in Fig.~\ref{fig:comparisonpromp} (b). As shown, ACNMP could accomplish the interpolation tasks already from the beginning and adapt to extrapolation tasks by sampling around 130 roll-outs. In comparison, \cite{Stark2019} achieved the task using around 1200 roll-outs where the majority of the cases (80\%) were from interpolation. In short, ACNMP does not require further adaptation as in \cite{Stark2019} for interpolation cases, and it achieves superior performance especially for extrapolation cases.

\begin{wraptable}{r}{4.5cm}
\centering
\vspace{-0.5cm}
\caption{Comparison summary}
\vspace{-0.2cm}
\label{sim_comparison}
\scalebox{0.68}{
\begin{tabular}{|c|c|c|}
\hline
 & ACNMP & ProMP+RL \\ \hline
\# of traj. &130 & 1200 \citep{Stark2019}\\  \hline
\makecell{\# of traj. \\(single env.)} &1.4K & 20K \citep{Ewerton2019}\\  \hline
\makecell{\# of traj. \\(total)} &0.7M & 10M \citep{Ewerton2019}\\  \hline
\end{tabular}
}
\vspace{-0.35cm}
\end{wraptable}
In the second experiment, we replicated the 2D obstacle avoidance experiment \citep{avoid} of \citet{Ewerton2019}, which optimizes trajectories based on the concept of relevant and generates trajectories using given task parameters via RL. In this experiment, we show that demonstrations for ACNMP do not need to be aligned temporally, please refer to the Appendix \ref{non-alignment} for details. As shown in Fig.~\ref{fig:comparisonpromp} (c) the task of the system is to generate obstacle avoidance trajectories that reach given goal points. The task parameters are composed of coordinates of the center of the hole in wall$(x_c, y_c)$ and coordinates of the goal point $(x_g, y_g)$, and RL minimizes the distance to the start point, the distance to the goal point and the signed distance to the walls in both  \citep{Ewerton2019} and our ACNMP model (see the Appendix \ref{non-alignment} for details.). In order to create training and test environments, these coordinates are sampled from a uniform distribution from the range of $2 \leq x_c \leq 8$, $1 \leq y_c \leq 9$, $x_c + 1.5 \leq x_g \leq 10$ and $0 \leq y_g \leq 10$. Both models are initialized with 30 random environments. Next, the models use a self-improvement loop via RL to increase the performance gradually. The performance of both methods is evaluated using a distinct test set composed of 1000 environments and figures~\ref{fig:comparisonpromp} (d) and~\ref{fig:comparisonpromp} (e) provides the change in error with respect to the number of evaluated environments and trajectories, respectively. As shown, while both methods are competitive in terms of the number of environments used, ACNMP outperforms the \cite{Ewerton2019} when the number of sampled trajectories are considered. Because ACNMP can learn the avoidance task for each environment from 1400 new trajectories maximum, their method required 20,000 new trajectories. In summary, the performance achieved by ACNMP with 700K trajectory samples could be reached by their method using 10 Million samples.In the table \ref{sim_comparison}, ACNMP's performance is summarized in comparison with \citep{Stark2019} and \cite{Ewerton2019}. For each environment and task, ACNMP requires fewer samples due to model's high representation capabilities and continuous exploitation of learned representations during RL. Note that our model optimizes more parameters as it uses neural networks (CNP \cite{CNP}) at the core, which has a higher computational requirement.
\vspace{-.2cm}
\begin{figure}[h]
    \centering
    \includegraphics[width=0.82\linewidth]{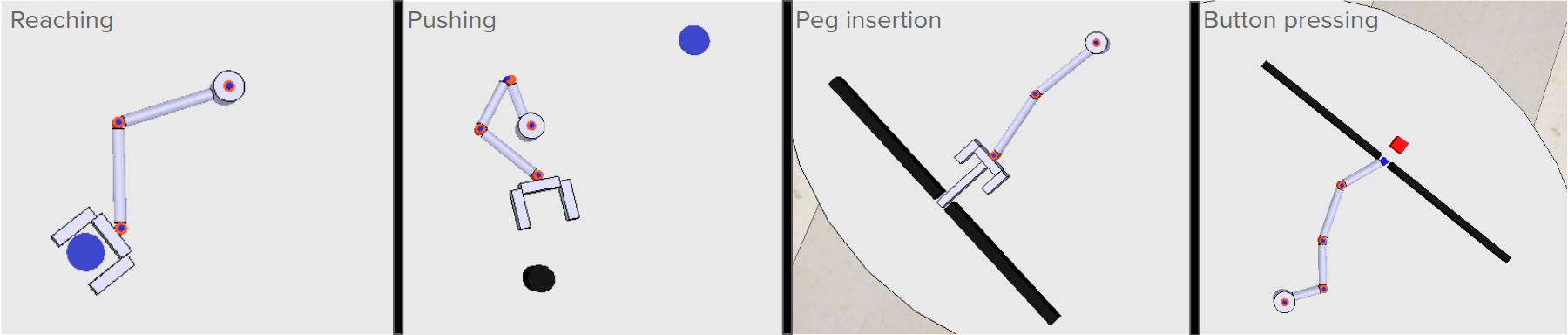}
    \caption{The three training tasks (on the left) and the test task (on the right) for transfer learning.}\vspace{-.2cm}
    \label{fig:transferexp}
\end{figure}
\vspace{-.3cm}
\begin{figure}[h]
\centering
\includegraphics[width=0.85\linewidth]{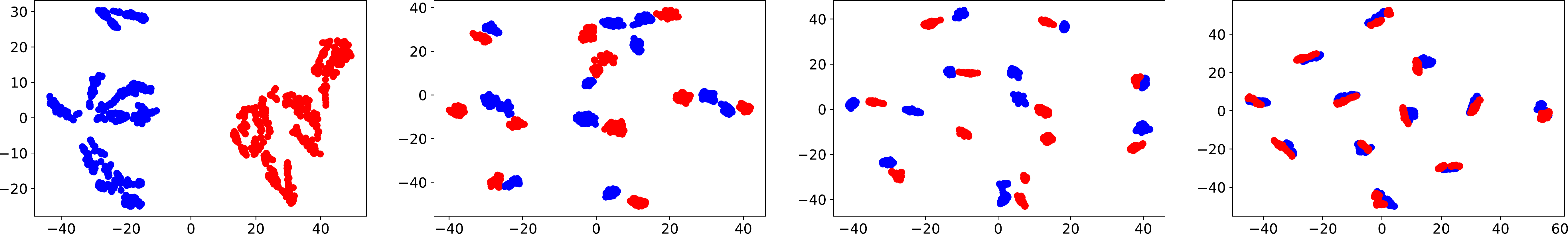}
\caption{Snapshots of overlaid latent space representations  obtained during learning.}
\label{fig:acnmptransferlatent}
\vspace{-.6cm}
\end{figure}

\subsection{Skill Transfer between Robots with Different Morphologies}
\vspace{-.2cm}
This experiment aims to evaluate the transfer learning capability of ACNMP, where two robots with different morphology are provided with the demonstration of a proxy skill.  Additionally,  demonstrations from a new task is made available only to the first robot. Then, we evaluate how fast the second robot can learn the new task by exploiting the behaviour of the first robot on the new task. For this, two models of the robots are trained for the proxy skill while forcing the latent representation in each robot's network to be similar. We adopted the experiment setup in \citep{gupta2017learning} for transfer learning. The two (3- and 4-dof) planar robot arms are simulated in CoppeliaSim~\citep{coppeliaSim} using the PyRep toolkit~\citep{james2019pyrep}. Both robots are provided with four different trajectories for reaching, pushing and peg insertion as shown in Fig.~\ref{fig:transferexp}. First, our system learns these skills and forms similar latent representations in the two robots (see Fig.~\ref{fig:acnmptransferlatent}).
The 3-dof robot is further provided with new demonstrations for a new task, namely button pressing task, where the robot must pass through a narrow opening and push the blue cube to the position of the red cube (Fig.~\ref{fig:transferexp}). The latent space activation is transferred to the 4-dof robot, which initially generated a trajectory close to the required one, but could not achieve the task. Via the same sparse reward used in \cite{gupta2017learning}, the distance between the two cubes, our ACNMP model was able to achieve 23, 73 and 90 percent success rate in 3, 7 and 11 iterations respectively and reached perfect success rate in 19 iterations.
While both methods show similar performances, \cite{gupta2017learning} obtained these results through explicitly learning a pairing of states observed in two robots and but they required an explicit time-warping mechanism. In our case, importantly, no time-warping and no explicit mechanism that ensures state-to-state match were required; our method learned the correspondences in trajectory level. 
\vspace{-.2cm}
\subsection{Extrapolation in Real-Robot Adaptation}
\begin{figure}[t]
    \includegraphics[width=1\linewidth]{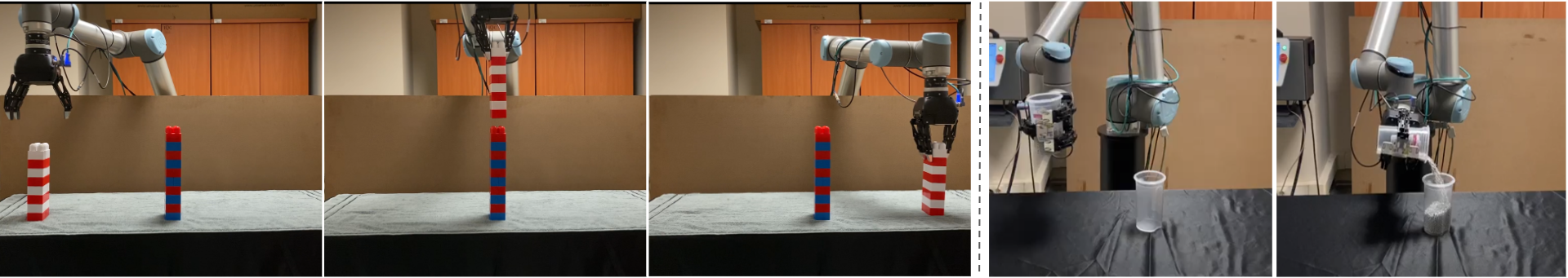} 
    \caption{Left: 3 snapshots from the pick \& place task. Right: 2 snapshots from the pouring task.}\vspace{-.2cm}
    \label{fig:realrobot}
\end{figure}

\vspace{-.1cm}
The first real robot experiment \cite{cnmpgithub} aims to verify the ACNMP in extrapolating to conditions in an object pick and place task that requires obstacle avoidance using a 6-dof UR10 robotic arm mounted with a Robotiq gripper (Fig.~\ref{fig:realrobot}). LfD-only model failed to generate suitable trajectories in configurations with objects and obstacles higher than the demonstrated ones \cite{sekerconditional}. Similar to \cite{sekerconditional}, ACNMP was provided with 8 demonstration trajectories for object and obstacle configurations from the range of [2-6] cm and [2-8] cm, respectively. The task parameters are set as the heights of the objects and the reward is set as the sum of distances to the grasp and obstacle avoidance points. ACNMP was able to adapt to novel task parameters (i.e. object and obstacle heights) that are non-linearly related to the required trajectories (Table.~\ref{tab:realresults} left). 

\begin{table}
\vspace{-0.2cm}
\caption{\label{tab:realresults} Left: Average joint errors in pick \& place task. Right: Error change in pouring during RL. }
\begin{minipage}[b]{0.67\linewidth}\centering

\resizebox{\textwidth}{!}{\begin{tabular}{|c|c|c|c|c|c|c|c|} \hline
\multicolumn{2}{|c|}{} & \multicolumn{6}{c|}{\bf Joint Errors (deg)} \\ \hline

\bf Method  &\bf Parameters  & \bf Base & \bf Shoulder & \bf Elbow  & \bf Wrist1  & \bf Wrist2 & \bf Wrist3 \\  \hline \hline
CNMP & Obj: $8^*$ Obs.: $10^*$ & $1.038$  & $2.325$ & $8.879$ & $8.094$ & $0.360$ & $0.830$  \\ \hline
ACNMP & Obj: $8^*$ Obs.: $10^*$ & $0.889$  & $1.589$ & $4.668$ & $3.465$ & $0.845$ & $0.861$  \\ \hline \hline
CNMP & Obj: $9^*$ Obs: $11^*$ & $1.861$  & $5.380$ & $17.978$ & $14.894$ & $0.590$ & $1.386$  \\ \hline
ACNMP & Obj: $9^*$ Obs: $11^*$ & $0.994$  & $0.507$ & $1.780$ & $1.514$ & $0.814$ & $1.007$  \\ \hline
\end{tabular}}
\end{minipage}
\begin{minipage}[b]{0.35\linewidth}\centering
\resizebox{.8\textwidth}{!}{
\begin{tabular}{|c|c|c|c|c|}
\hline
     & \multicolumn{4}{c|}{\bf Error in Grams}  \\ \hline
1-4    & 591 & 447 & 435 & 175 \\ \hline
5-8    & 511 & 604 & 305 & 42  \\ \hline
9-12   & 417 & 284 & 144 & 194 \\ \hline
13-16  & 367 & 364 & 79  & 177 \\ \hline
17-18  & 133 &  \textbf{27} & &  \\ \hline
\end{tabular}
}
\end{minipage}
\vspace{-0.5cm}
\end{table}

Above, the extrapolation capability of the ACNMP is verified when it is conditioned on novel points in its movement trajectory. 
The aim of the second real robot experiment is to show that ACNMPs can also adapt when conditioned on task parameters that are outside the range of training and using the reward coming from the real-world. The robot is given a cup that contains different amounts of marbles with the aim of pouring the target amount of marbles into another cup (Fig.~\ref{fig:realrobot} right). Depending on the weight of the cup and the target pouring amount, the robot is expected to generate wrist joint trajectories with suitable amount of rotation. Therefore, ACNMPs need to learn to generate joint trajectories conditioned with the initial weight of the cup and desired pouring amount.
16 expert joint trajectories were given with initial cup weights in the range of 650-1200 grams and rotation angles in the range of 85-95 degrees. Given the extrapolation parameter, 1400 gram initial cup weight and 1322 grams target pouring amount, ACNMP found an RL solution and assimilated it into the primitive after 18 roll-outs. While the LfD approach showed 600 gram error on this extrapolation task, the error rate of ACNMP dropped to 27 grams after training (Table~\ref{tab:realresults} right). In real robot experiments, we used an explicit controller to drive the robot since ACNMP produces trajectories.
\vspace{-.2cm}
\section{Conclusion}
\label{sec:conclusion}
\vspace{-.2cm}
In this paper, we proposed a new adaptive movement primitive approach, namely ACNMPs, by integrating LfD and RL to address the challenges of efficient sampling, skill maintenance and skill transfer.
ACNMPs have been shown to outperform two other adaptive movement primitive learning approaches in solving extrapolation tasks in terms of adaptation performance and data-efficiency while still retaining the qualitative characteristics. ACNMPs were shown to learn meaningful latent feature representations from the data, thus contributing to a successful generalization. As a step further, we exploited the latent representation to perform domain transfer between robots with different morphologies to solve an unknown task in a few-shot learning manner. Finally, the applicability of the ACNMPs to real-world scenarios was demonstrated by real robot experiments involving obstacle avoidance in reaching and manipulation tasks. In the future, we would like to seek mathematical investigations and theoretical guarantees on shape preservation and assess our method on tasks including rich contacts with real robots.  

\clearpage
\acknowledgments{This research has received funding from the European Union's Horizon 2020 research and innovation programme under grant agreement no. 731761, IMAGINE. It was partially supported by JST CREST “Cognitive Mirroring” [grant number: JPMJCR16E2], by the International Joint Research Promotion Program, Osaka University and TUBITAK
(Scientific and Technological Research Council of Turkey)
2210-A scholarship. We would like to thank Yukie Nagai, Svenja Stark and Elmar Rueckert for their comments and feedback on this work and the reviewers for their suggestions.}

\bibliography{references}  
\appendix

\section{Implementation Details}
\label{implementation_details}
For encoder and decoder model, neural networks with ReLU activations were used and they were trained using Adam optimizer \citep{kingma2014adam}. In addition to given source codes, network details can be found in Table \ref{table:net_details} for full reproducibility. Note that our focus was not on optimizing network sizes even though we observed that similar accuracies can be obtained using smaller networks. For reinforcement learning, we used Retrace algorithm \citep{munos2016safe} and in the transfer learning task, latent space visualization (Figure \ref{fig:acnmptransferlatent}) was obtained using T-SNE method \citep{smaaten2008visualizing}.

\begin{table}[h]
\caption{Neural network details for different tasks. Encoder and decoder structure show number of neurons for each layer. 
}

\centering
\begin{tabular}{l|l|l}
                                       & Encoder Structure & Decoder Structure \\ \hline
Extrapolation in 2D Obstacle Avoidance (4.1) & 128,64,32,16,8 &  124,124,2  \\ \hline
Extrapolation in 2D Pushing Task (4.2)    &   128,128,64,32 &  128,128,128,6                 \\ \hline
Wall Avoidance (4.2)                        &  128,128,128,64 & 128,128,64,32,4   \\ \hline
Transfer Learning, Model of 3 Dof (4.3)     &  128,128,128,64 & 128,128,128,128,6                 \\ \hline
Transfer Learning, Model of 4 Dof (4.3)     &  128,128,128,64 & 128,128,128,128,8                  \\ \hline
Pick and Place (4.4)                       &   128,128,64,32 &   128,128,128,12                \\ \hline
Water Pouring (4.4)                          & 128,128,64,32 & 128,128,128,2     \\ \hline
\end{tabular}

 \label{table:net_details}
\end{table}

\section{Further analysis of extrapolation amount} 
\label{extreme_extrapolation}
\begin{figure}[h]
\centering
    \includegraphics[width=0.8\linewidth]{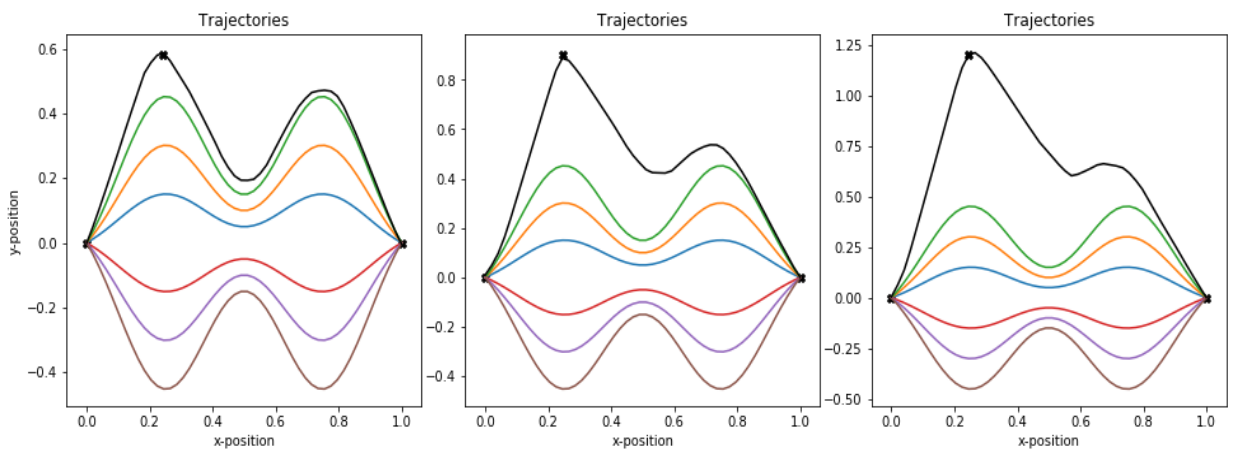}
    \caption{ACNMP performance in response to the increasing amount of extrapolation.
    Colored and black lines correspond to the demonstrated and generated trajectories, respectively.
    }
    \label{fig:limitations}
\end{figure}

Recall that in section 4.1, we showed performance of ACNMP for extrapolation point in 2D obstacle avoidance task. 
In this analysis, we investigated to what extent our algorithm is successful by placing the extrapolation point further away from the demonstrations. Although the algorithm still attempts to find solutions that satisfy the constraints, the shapes of the generated trajectories change significantly (see Fig.~\ref{fig:limitations}). Note that given only one constraint, ``what must the right shape of the complete trajectory be?'' is not a well-defined question.

\section{Demonstrations for wall avoidance experiment (4.3)}
\label{non-alignment}
\begin{figure}[h]
    \includegraphics[width=\linewidth]{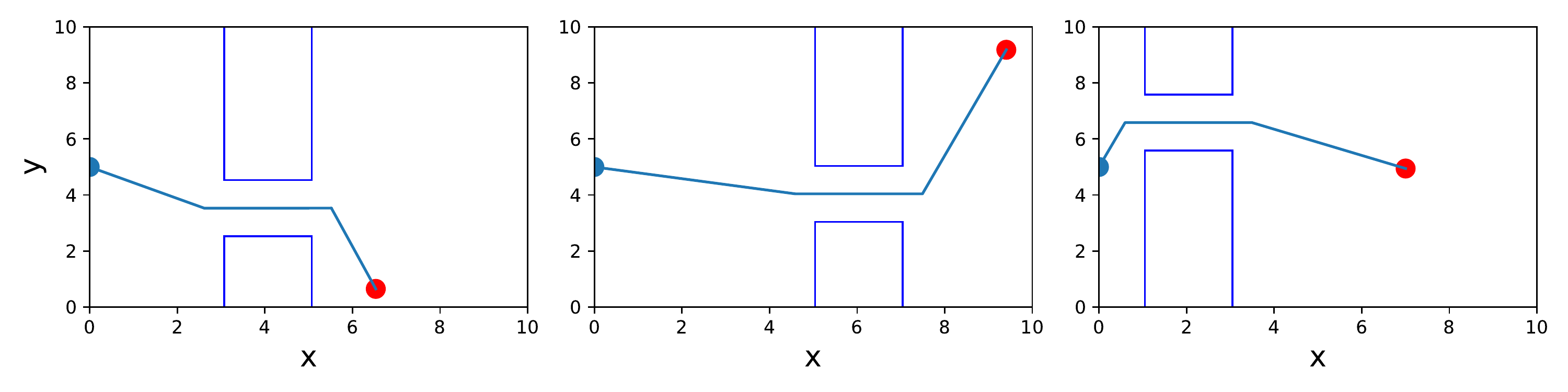}
    \caption{The demonstrations for the wall avoidance experiment are shown. The blue and red circles show the start and end points respectively.
    }
    \label{fig:wall_avoidance_demos}
\end{figure}

\textbf{Temporal alignment}:
Recall that in section 4.3, we showed ACNMP's performance in a 2D simulation wall avoidance experiment where the position of the wall and goal point is changing. Importantly, in this experiment, we verify that the demonstrations do not need to be time-aligned. In figure \ref{fig:wall_avoidance_demos}, the demonstrations are illustrated. Note that the trajectory is traveled uniformly in time and avoiding the wall behavior should occur at different time steps.
\textbf{Reward function}:
\cite{Ewerton2019} stated that the third component of reward was set as the minimum distance to the center of the hole but plots show different objective. After personal contact with the authors, the exact reward was clarified as stated in the text.

\section{Joint trajectories for the first real robot experiment}
\begin{figure}[h]
\centering
\includegraphics[width=0.9\linewidth]{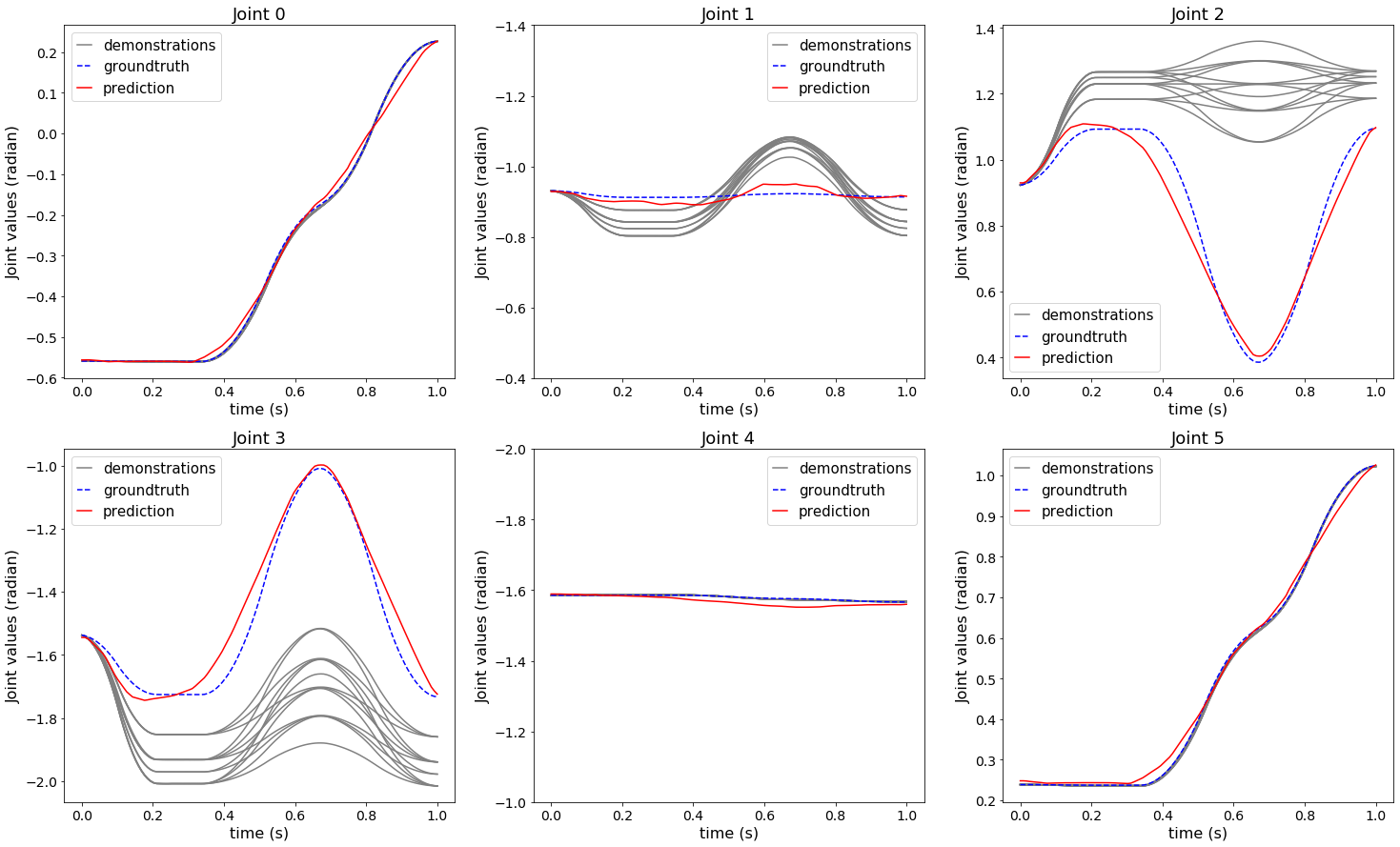}
\caption{ACNMP predictions for pick and place task. Expert demonstrations are shown in gray, corresponding expert trajectories for extrapolation task parameters are shown in blue and ACNMP predictions are shown in red.  
}
\label{fig:sixjoints}
\end{figure}

Recall that in section 4.4, we showed performance of ACNMP for real world pick and place task, where six joints of UR10 robot were optimized to carry an object while avoiding an obstacle. Here, we would like to provide ACNMP joint trajectories for pick and place real robot task in figure \ref{fig:sixjoints}.  It can be seen that the found solution is similar to the expert solution in terms of shape, even though there is no reward for similarity. 

\end{document}